\documentclass[10pt,twocolumn,letterpaper]{article}

\usepackage{wacv}
\usepackage{times}
\usepackage{color}
\usepackage{epsfig}
\usepackage{graphicx}
\usepackage{amsmath}
\usepackage{amssymb}
\usepackage{subfigure}
\usepackage{subcaption}

\usepackage[pagebackref=true,breaklinks=true,colorlinks,bookmarks=false]{hyperref}

\wacvfinalcopy 

\def\wacvPaperID{961} 

\ifwacvfinal\fi
\begin{document}

\title{Assessing Cardiomegaly in Dogs Using a Simple CNN Model}

\author{Nikhil Deekonda \\
Yeshiva University, NYC, NY\\
{\tt\small ndeekond@mail.yu.edu}
}

\maketitle
\ifwacvfinal\thispagestyle{empty}\fi

\begin{abstract}
This paper introduces DogHeart\cite{Li2024}, a dataset comprising 1400 training, 200 validation, and 400 test images categorized as small, normal, and large based on VHS score. A custom CNN model is developed, featuring a straightforward architecture with 4 convolutional layers and 4 fully connected layers. Despite the absence of data augmentation, the model achieves a 72\% accuracy in classifying cardiomegaly severity. The study contributes to automated assessment of cardiac conditions in dogs, highlighting the potential for early detection and intervention in veterinary care.

\end{abstract}


\section{Introduction}
Cardiac disease is a prevalent health concern among dogs, with cardiomegaly, the abnormal enlargement of the heart, being a significant indicator of various underlying cardiac conditions. Early detection and accurate assessment of cardiomegaly are crucial for effective treatment and management, potentially extending the lives of affected animals. Traditional diagnostic methods rely heavily on manual interpretation of radiographs by veterinarians, which can be time-consuming and subject to variability in expertise.

In recent years, deep learning techniques have demonstrated considerable promise in the field of medical image analysis, offering automated and consistent diagnostic capabilities. With the development of convolutional neural networks (CNN), radiologists autonomously identify complicated patterns with computer vision algorithms that are accurate for all imaging modalities. Since most degenerative canine heart diseases accompany cardiomegaly, early detection of cardiac enlargement is a priority healthcare issue for dogs\cite{jeong2022automated}. Applying AI technologies to dog cardiomegaly assessment can not only reduce the time and costs involved in pet diseases diagnosis and treatment, but also expand their use in the less AI-focused veterinary medicine field, compared to human medicine.\cite{oh2023leveraging}

This work seeks to bridge the gap between advanced deep learning methodologies and practical veterinary applications, offering a robust solution for the early detection and classification of cardiomegaly. By providing an automated and reliable assessment tool, I hope to assist veterinary professionals in making more informed decisions, ultimately contributing to better health outcomes for canine patients.

\section{Related Work}\label{sec:related}

Deep learning techniques have recently been introduced to aid the VHS (vertebral heart size) method in diagnosing canine cardiomegaly in veterinary medicine. Zhang et al.\cite{zhang2021computerized} utilized the positions of 16 key points detected by a deep learning model to calculate the VHS value, which was then compared with the VHS reference range for all dog breeds to evaluate canine cardiomegaly. Jeong and Sung\cite{jeong2022automated} introduced the "adjusted heart volume index" (aHVI), a new deep learning-based radiographic index, using retrospective data to quantify canine heart size for diagnostic purposes. Burti et al.\cite{burti2020use} developed a convolutional neural network (CNN)-based computer-aided detection (CAD) system to detect cardiomegaly from plain radiographs in dogs. Dumortier et al.\cite{dumortier2022deep} employed a ResNet50V2-based CNN to classify feline thoracic radiograph images, distinguishing between cats with and without Radiographic Pulmonary Patterns (RPPs), and proposed an optimized framework for improved performance. Müller et al.\cite{muller2022accuracy} created an AI algorithm to identify pleural effusion in thoracic radiographs of dogs. Zhang and Li\cite{Li2024} proposed a regressive vision transformer (RVT) model for dog cardiomegaly classification, which is not limited to radiograph X-ray image diagnosis, but can be applied to other types of medical images, such as CT scans and ultrasounds. Their model can be extended to detect human cardiomegaly using different diagnosis technologies.

\section{Methods}\label{sec:method}
\subsection{Motivation}
This study primarily aims to develop a neural network specialized in accurately classifying the severity of cardiomegaly in dogs. The goal is to achieve a classification accuracy of at least 70\% on test data, thereby providing a reliable tool for veterinary professionals to diagnose and manage this critical condition. Additionally, the study seeks to explore the effectiveness of a simplified CNN architecture in achieving high classification accuracy. By providing an automated and efficient assessment method, this research aspires to enhance early detection and intervention, ultimately improving the health and well-being of canine patients.
\subsection{Environment Setup}
To ensure the effectiveness of our approach, we employed PyTorch version 2.0.1 in conjunction with torchvision version 0.15.2. This framework was complemented by CUDA version 12.2, utilizing the parallel processing capabilities of the NVIDIA GPU. Our model training was conducted on Google Colab, a cloud-based platform providing access to high-performance computing resources, including an NVIDIA Tesla T4 GPU.

Given the computational intensity of the training tasks, all training procedures were performed exclusively on the GPU. This approach was intended to leverage the GPU's accelerated processing capabilities, ensuring optimal performance during the training phase. The choice of this hardware configuration facilitated efficient computation and resource utilization, thereby enhancing the overall quality and efficiency of our approach.
\subsection{Data Preprocessing}

Each image was resized to $75 \times 75$ pixels and converted to tensor format using the following transformations:
\begin{itemize}
    \item \textbf{Resize}: Each image was resized to $75 \times 75$ pixels to ensure uniform input dimensions.
    \item \textbf{ToTensor}: The images were then converted to tensors for compatibility with PyTorch models.
\end{itemize}

\subsection{Model Architecture}

\begin{figure*}[h]
    \centering
    \includegraphics[width=\textwidth]{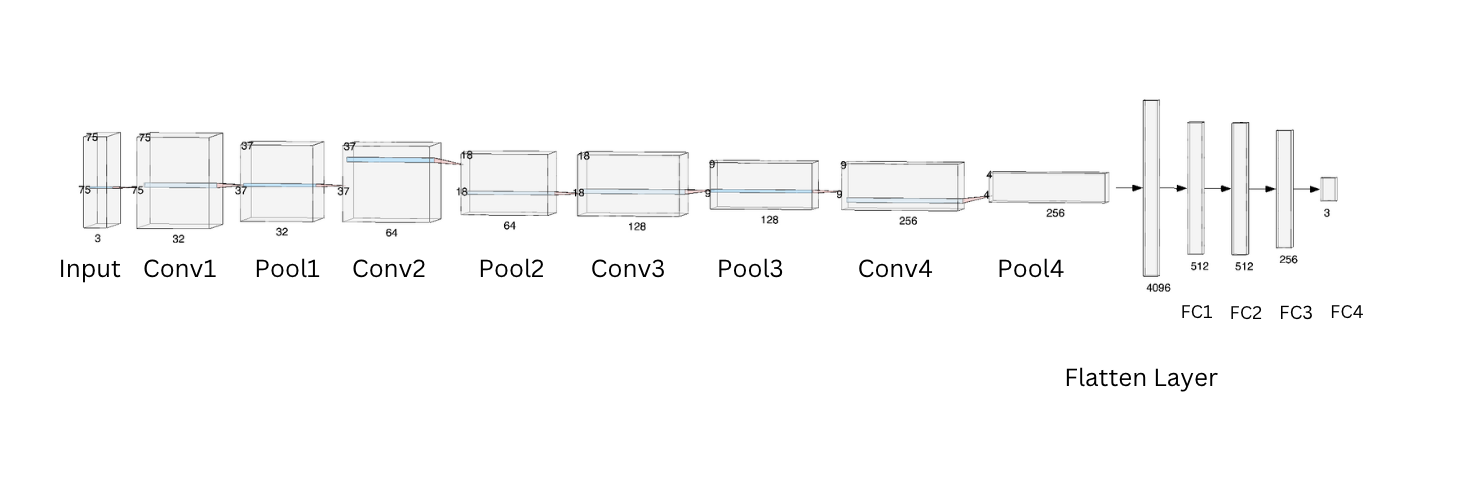}
    \caption{Architecture of the custom CNN model.}
    \label{fig:cnn_architecture}
\end{figure*}

The custom convolutional neural network (CNN) architecture employed in this study consists of four convolutional layers followed by four fully connected layers. The architecture is illustrated in Figure \ref{fig:cnn_architecture}.

\subsection{Training and Hyperparameters}

The training process was carried out using the Adam optimizer with a learning rate of 0.001 and the cross-entropy loss function. The batch size was set to 32, and the model was trained for 50 epochs. The following hyperparameters were used:
\begin{itemize}
    \item \textbf{Optimizer}: The choice of optimizer influences how the model's parameters are updated. We employed the Adam optimizer, known for its adaptive learning rate and momentum capabilities.
    \item \textbf{Learning Rate}: 0.001, the rate at which the model updates its parameters during optimization.
    \item \textbf{Batch Size}: 32, this determines the number of samples processed before the model's internal parameters are updated
    \item \textbf{Epochs}: 50, this specifies the number of iterations over the entire dataset during the training process.
\end{itemize}
\subsection{Learning Rate and Optimization Algorithm}

Hyperparameters such as the learning rate and the choice of optimizer are crucial in determining the convergence behavior and final performance of the model. In this project, we utilized the Adam optimizer, known for its adaptive learning rate and efficient handling of sparse gradients, making it well-suited for training deep neural networks.

The learning rate was initially set to $0.001$, a value selected based on empirical testing and consideration of the dataset characteristics and model architecture. The Adam optimizer incorporates both the advantages of AdaGrad and RMSProp, adjusting the learning rate for each parameter dynamically.

The parameter update rule for the Adam optimizer is given by:

\begin{align}
m_t &= \beta_1 m_{t-1} + (1 - \beta_1) \cdot g_t \\
v_t &= \beta_2 v_{t-1} + (1 - \beta_2) \cdot g_t^2 \\
\hat{m}_t &= \frac{m_t}{1 - \beta_1^t} \\
\hat{v}_t &= \frac{v_t}{1 - \beta_2^t} \\
\theta_{t+1} &= \theta_t - \eta \cdot \frac{\hat{m}_t}{\sqrt{\hat{v}_t} + \epsilon}
\end{align}

where:
\begin{itemize}
    \item $m_t$ and $v_t$ are the first and second moment estimates, respectively,
    \item $g_t$ represents the gradient of the loss function with respect to the parameters at time step $t$,
    \item $\beta_1$ and $\beta_2$ are the exponential decay rates for the moment estimates,
    \item $\eta$ denotes the learning rate,
    \item $\epsilon$ is a small constant to prevent division by zero,
    \item $\theta_t$ represents the parameters at time step $t$.
\end{itemize}

Using the Adam optimizer with these settings, the model was able to converge efficiently, balancing between stability and adaptability in the parameter updates. This approach facilitated effective learning and contributed to achieving satisfactory performance on the validation dataset.

\subsection{Validation and Best Model Selection}

During training, the model's performance was evaluated on the validation dataset at the end of each epoch. The loss was computed as Train Loss and Validation Loss using cross entropy loss eq.~\ref{eq:cross}
\begin{equation}\label{eq:cross}
\text{Cross Entropy Loss} = -\frac{1}{N} \sum_{i=1}^{N} \sum_{j=1}^{C} y_{i,j} \log(p_{i,j})
\end{equation}

N is the number of samples in the batch, C is the number of classes,
$y_{i,j}$ is the ground truth label for sample i and class j, represented as a one-hot encoded vector, and $p_{i,j}$ is the predicted probability of sample i belonging to class j.

The model that achieved the lowest validation loss during the training process was selected as the best model. This model was then saved for further evaluation on the test dataset. The training and validation loss were monitored and reported at the end of each epoch to ensure that the model was not overfitting.

\subsection{Test Dataset Evaluation}

The test dataset, consisting of 400 images, was used to evaluate the final model's performance. Since the test dataset labels were not available, predictions were generated using the trained model. These predictions were then evaluated using software provided in \href{https://github.com/YoushanZhang/Dog-Cardiomegaly}{Zhang's GitHub repository}, which computes accuracy scores based on the predicted results.

\section{Results}\label{sec:results}
\subsection{Datasets}

The dataset used in this study, DogHeart, comprises a total of 2,000 digital images, divided into 1,400 training images, 200 validation images, and 400 test images. Each image is categorized into one of three classes: small, normal, and large, based on the severity of cardiomegaly. Table \ref{tab:dataset_distribution} provides a detailed breakdown of the dataset distribution across these classes.

\begin{table}[h]
    \centering
    \caption{Dataset Distribution}
    \label{tab:dataset_distribution}
    \begin{tabular}{lccc}
        \hline
        \textbf{Dataset} & \textbf{Large} & \textbf{Normal} & \textbf{Small} \\
        \hline
        Training & 619 & 573 & 208 \\
        Validation & 76 & 91 & 33 \\
        Test & \multicolumn{3}{c}{400 (labels not available)} \\
        \hline
    \end{tabular}
\end{table}

\begin{figure*}[!htbp]
    \subfigure[Large]{\label{fig:large}
        \includegraphics[width=0.2\linewidth]{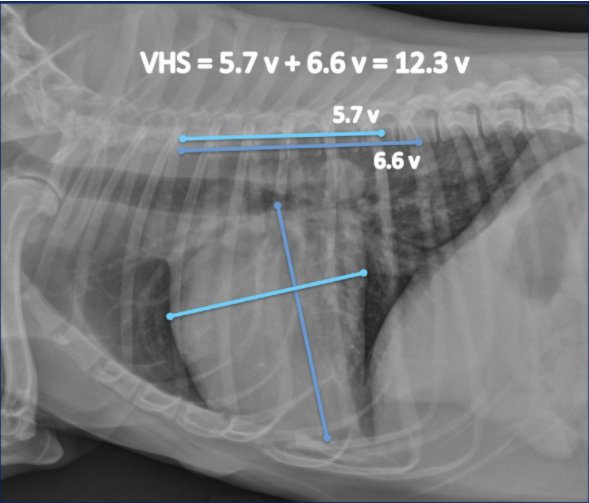}}%
    \hspace{\fill}
    \subfigure[Normal]{
        \includegraphics[width=0.2\linewidth]{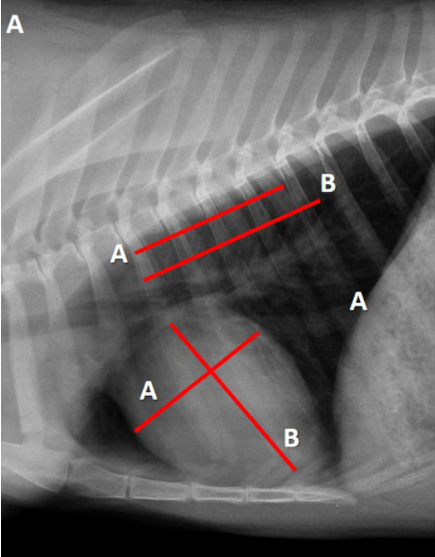}\label{fig:normal}}%
    \hspace{\fill}
    \subfigure[Small]{
        \includegraphics[width=0.2\linewidth]{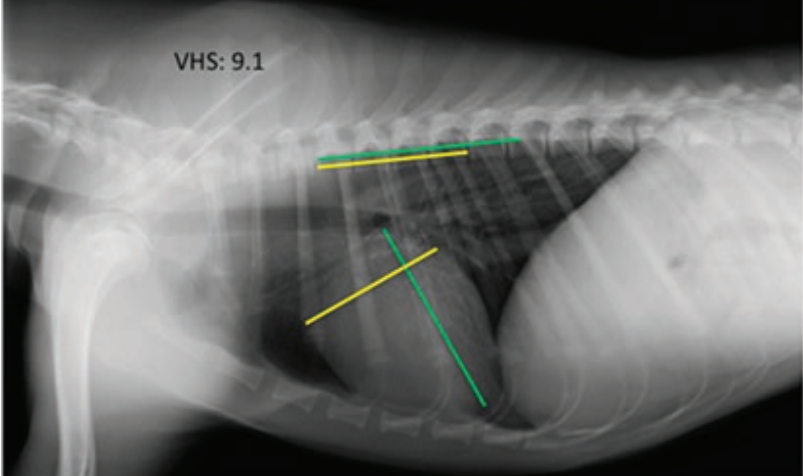}\label{fig:small}}
    \caption{Example images of each class}
    \label{fig:example_images}
\end{figure*}

The images were collected from \href{https://github.com/YoushanZhang/Dog-Cardiomegaly}{Zhang Github Repository}. Each image corresponds to an individual dog. All images with vertebral heart scale (VHS) scores below 8.2 are classified as small hearts, normal dogs are between 8.2 and 10, and large dogs are above 10. Table \ref{tab:dataset_distribution} show that there are fewer samples of the small dog category, and the number of normal and large dog categories are balanced in the DogHeart dataset. Figure \ref{fig:example_images} presents typical examples of each class, showcasing the variability in heart sizes.

\subsection{Results}
During training, the model consistently improved its performance, achieving a validation loss of 0.5826. This metric reflects the model's ability to accurately classify the severity of cardiomegaly into three categories: small, normal, and large. The model achieved an accuracy of 72\% on the test dataset, demonstrating the effectiveness of the chosen architecture. 



\section{Challenges and Limitations}

\subsection{Data Imbalance}
One of the primary challenges faced in this project is the imbalance in the dataset classes. The training dataset contains significantly fewer images for the 'Small' class compared to the 'Large' and 'Normal' classes. This imbalance can lead to the model being biased towards the more frequently occurring classes, potentially affecting its performance in accurately predicting the 'Small' class.

\subsection{Computational Resources}
Training deep learning models, especially custom CNN architectures, requires substantial computational power. The resources available for this project, while sufficient, may not be optimal for exploring more complex architectures or performing extensive hyperparameter tuning. This limitation might restrict the exploration of more advanced models that could potentially improve accuracy.

\subsection{Model Complexity}
The custom CNN model, although simpler and more computationally efficient, might lack the depth and complexity required to capture more intricate patterns in the data compared to more sophisticated models like VGG16. This limitation can potentially cap the performance of the model, preventing it from achieving state-of-the-art results.

\subsection{Generalization}
The model's performance on the validation dataset is a good indicator of its potential real-world application, but there's always a risk that the model might not generalize well to unseen data. Overfitting to the training data can be a concern, especially with a relatively small dataset, and ensuring robust generalization is a persistent challenge.

\subsection{Evaluation Metrics}
While accuracy is a useful metric, it may not fully capture the model's performance, especially in the presence of class imbalance. Metrics such as precision, recall, F1-score, and confusion matrices are crucial for a more detailed understanding of the model's strengths and weaknesses across different classes.

\subsection{Data Augmentation}
The project does not incorporate data augmentation techniques, which could help in improving the model's performance by artificially increasing the size and variability of the training dataset. Techniques such as random cropping, flipping, rotation, and scaling could provide the model with a more diverse set of training examples, potentially leading to better generalization.

\section{Discussion}\label{sec:dis}
The custom model achieves an accuracy of 72\%, which exceeds the minimum requirement of 70\%. This demonstrates that the custom model is capable of performing the task to a satisfactory level. While VGG16 has a slightly higher accuracy (74.8\%), the difference is relatively small (2.8\% higher than the custom CNN model). This indicates that the custom model is competitive with the VGG16 architecture. Moreover, VGG16 is a very deep model with 16 layers, leading to a higher parameter count and potentially longer training times while our model has a simpler architecture with fewer layers, making it more computationally efficient. This model is likely faster to train and less prone to overfitting due to fewer parameters. In cases where computational resources or time are constraints, the custom model provides a more efficient alternative with comparable performance. With fine-tuning and optimization (e.g., adjusting learning rates, experimenting with data augmentation, or adding more regularization), the custom model might be able to narrow the performance gap further with VGG16.

\section{Conclusion}\label{sec:conclusion}

This custom model meets the minimum requirement with a solid accuracy of 72\%. Although VGG16 achieves a slightly higher accuracy of 74.8\%, my model's performance remains competitive. Additionally, the custom model offers advantages in terms of simplicity and computational efficiency, making it a viable choice for scenarios where resources are constrained. With further optimization, this model has the potential to achieve even better performance.

\end{document}